\documentclass{article}

\usepackage{microtype}
\usepackage{graphicx}
\usepackage{subfigure}
\usepackage{booktabs} 
\usepackage[english]{babel}
\usepackage[utf8x]{inputenc}
\usepackage[colorinlistoftodos]{todonotes}
\usepackage{amsmath}
\usepackage{amsfonts}
\usepackage{siunitx}
\usepackage{etoolbox}

\usepackage{hyperref}

\newcommand{\tinypm}{\raisebox{1.0pt}{$_{^\pm}$}}
\newcommand{\ubold}{\fontseries{b}\selectfont}
\robustify\ubold
\newcommand{\nri}{\textsc{nri} }

\usepackage[accepted]{icml2019}

\icmltitlerunning{Factorised Neural Relational Inference}

\begin{document}

\twocolumn[
\icmltitle{Factorised Neural Relational Inference for Multi-Interaction Systems}



\icmlsetsymbol{equal}{*}

\begin{icmlauthorlist}
\icmlauthor{Ezra Webb}{phys,equal}
\icmlauthor{Ben Day}{comp,equal}
\icmlauthor{Helena Andres-Terre}{comp}
\icmlauthor{Pietro Li\'o}{comp}
\end{icmlauthorlist}

\icmlaffiliation{phys}{Department of Physics, The Cavendish Laboratory, University of Cambridge, UK.}
\icmlaffiliation{comp}{Department of Computer Science \& Technology, The Computer Laboratory, University of Cambridge, UK.}

\icmlcorrespondingauthor{Ben Day}{ben.day@cl.cam.ac.uk}

\icmlkeywords{Machine Learning, Graph Networks, GNN, NRI, ICML}

\vskip 0.3in
]



\printAffiliationsAndNotice{\icmlEqualContribution} 
\begin{abstract}
Many complex natural and cultural phenomena are well modelled by systems of simple interactions between particles. A number of architectures have been developed to articulate this kind of structure, both implicitly and explicitly. We consider an unsupervised explicit model, the \textsc{nri} model, and make a series of representational adaptations and physically motivated changes. Most notably we factorise the inferred latent interaction graph into a multiplex graph, allowing each layer to encode for a different interaction-type. This f\textsc{nri} model is smaller in size and significantly outperforms the original in both edge and trajectory prediction, establishing a new state-of-the-art. We also present a simplified variant of our model, which demonstrates the \textsc{nri}'s formulation as a variational auto-encoder is not necessary for good performance, and make an adaptation to the \textsc{nri}'s training routine, significantly improving its ability to model complex physical dynamical systems.
\end{abstract}

\section{Introduction \& Related Work}
There are interesting phenomena at every physical scale that are well described by dynamical systems of interacting particles. Thinking about things in this way has proven to be a valuable method for investigating the natural world. As we come to develop more intelligent systems to assist in our investigations, the ability to work within this framework will be of great use.

Many systems have been developed to model interactions implicitly with either single-layer fully connected graphs \cite{Sukhbaatar2016LearningBackpropagation,Guttenberg2016Permutation-equivariantPrediction,Santoro2017AReasoning,Watters2017VisualNetworks} or with attention-based control mechanisms \cite{Hoshen2017VAIN:Modeling,vanSteenkiste2018RelationalInteractions}. However, to the end of developing an investigative or theorising machine assistant, modelling the interaction graph \textit{explicitly} is more valuable than a high quality trajectory-reconstruction. The neural relational inference (\textsc{nri}) model, introduced by Kipf et al. \yrcite{Kipf2018NeuralSystemsb}, is an unsupervised neural network that learns to predict the interactions and dynamics of a system of objects from observational data alone. When provided with the trajectories of a system of interacting objects, the model infers an explicit interaction graph for these objects which it uses to predict the evolution of the system.

The \textsc{nri} model presents a strong foundation, answering key architectural questions and opening the door for further work dealing with explicit representations. In this work we identify two problems within the original formulation -- representational and experimental -- and in addressing these develop a model that significantly outperforms the original. We also present a variant of our model with greatly improved trajectory prediction that demonstrates the \textsc{nri} model's formulation as a variational auto-encoder (\textsc{vae}) is not necessary for good performance.

Specifically, in a system with multiple independent interactions, representing the interaction relationships as a single graph with many edge-types requires exponentially many types to accommodate all possible combinations of interactions. Critically, feedback in such a system is unable to distinguish partially correct and entirely incorrect predictions. In this work we adopt a multiplex structure wherein different interactions are \textit{factorised} into separate layer-graphs, greatly compressing the representation whilst also permitting better directed feedback and improved training. 

\subsection{NRI in brief}
\label{S:nri_overview}
We provide an outline of the \textsc{nri} architecture with formal definitions given only for those parts that we modify\footnote{An extended description with original diagrams is provided as supplementary material.}. We adopt the formalism and nomenclature of Kipf et al. \yrcite{Kipf2018NeuralSystemsb} throughout.

Most simply, the \textsc{nri} takes the form of a variational-autoencoder (\textsc{vae}): trajectories are encoded as a latent interaction graph that is decoded when predicting trajectories for given initial conditions. A trajectory is a series of features over time, where $\mathbf{x}^t_i$ is the feature vector of the $i$-th object at step $t$. The latent interaction graph has $K$-many edge-types encoded as one-hot vectors, where  $\mathbf{z}_{ij}$ is the edge-type vector between nodes (objects) $i$ and $j$.

\paragraph{Encoder}

The encoder receives each particle's trajectory as the feature of its corresponding node in a fully-connected graph and produces an edge-type vector for each pair of particles. A graph neural network (\textsc{gnn}) computes a series of message passing operations \cite{pmlr-v70-gilmer17a} and produces a $K$-dimensional edge-embedding vector $\mathbf{h}^2_{(i,j)}$ for each pair of particles $(i,j)$.\footnote{$\mathbf{h}^2_{(i,j)}$ is used to align with the original paper's notation.}

\paragraph{Posterior distribution}
The edge-type posterior distributions are taken as $q_{\theta}(\mathbf{z}_{ij}|\mathbf{x}) = \text{softmax}(\mathbf{h}^2_{(i,j)})$, from which the edge-type vectors $\mathbf{z}_{ij}$ are sampled, where $\theta$ summarizes the parameters of the full encoder \textsc{gnn}.

\paragraph{Decoder}

The task of the decoder is to predict the dynamics of the system using the latent interaction graph $\mathbf{z}$ and the past dynamics. We consider the Markovian case; calculating $p_{\phi}(\mathbf{x}^{t+1}|\mathbf{x}^t; \mathbf{z})$. The message passing section consists of
\begin{align}
\label{eq:intro_NRI_decoder1}
v \rightarrow e: \tilde{\mathbf{h}}^t_{(i,j)} &= \sum_{k=1}^K z_{ij,k} \tilde{f}^k_e \big( [\mathbf{x}_i^t,\mathbf{x}_j^t] \big) \\
\label{eq:intro_NRI_decoder2}
e \rightarrow v: \boldsymbol{\mu}_j^{t+1} &= \mathbf{x}_j^t + \tilde{f}_v \big( \big[ \textstyle\sum\nolimits_{i \neq j} \tilde{\mathbf{h}}^t_{(i,j)}, \mathbf{x}_j^t \big] \big)
\end{align}
where $[\cdot,\cdot]$ denotes concatenation. We note that each edge-type $k$ has its own function in the edge-to-vertex message passing operation -- $\tilde{f}^1_e,\dots,\tilde{f}^K_e$. The future state of each object is then sampled from an isotropic Gaussian distribution with fixed (user-defined) variance $\sigma^2$
\begin{align}
p_{\phi}(\mathbf{x}_j^{t+1}|\mathbf{x}^t,\mathbf{z}) = \mathcal{N}( \boldsymbol{\mu}_j^{t+1}, \sigma^2 \mathbf{I} ). \nonumber
\end{align}

\paragraph{Objective}
The model is trained as a \textsc{vae} maximising the evidence lower bound
\begin{align}
\label{eq:intro_ELBO}
\mathcal{L} =\ \ &\mathbb{E}_{q_{\theta}(\mathbf{z}|\mathbf{x})}[ \log p_{\phi}(\mathbf{x}|\mathbf{z}) ] - D_{\textsc{kl}}[ q_{\theta}(\mathbf{z}|\mathbf{x})||p(\mathbf{z}) ]
\end{align}
where $D_{\textsc{kl}}$ is the Kullback-Leibler (\textsc{kl}) divergence. It is also relevant to note that the reconstruction error is estimated by a re-scaled mean-squared error (\textsc{mse}) of $\boldsymbol{\mu}$ relative to $\mathbf{x}$.\footnote{Re-scaled by the hyperparameter $\tfrac{1}{2\sigma^2}$ (plus a constant).}

\section{Model}

\begin{figure*}[ht]
\centering
\includegraphics[width=15cm,trim=0cm 0cm 0cm 0.5cm]{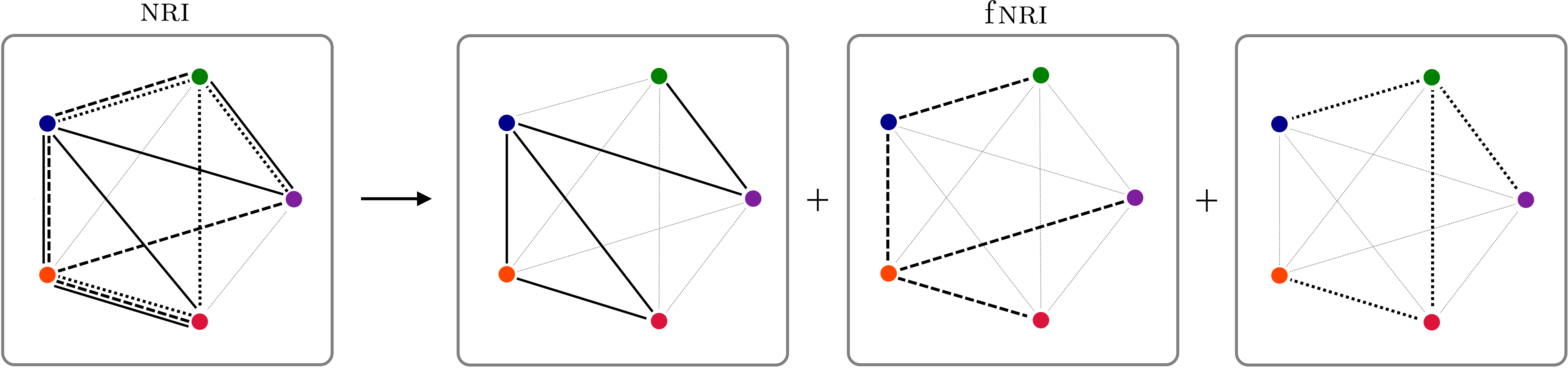}
\caption{\footnotesize Schematic showing the representational change in the interaction graph between the \textsc{nri} and f\textsc{nri} models when there are three independent interaction types represented by solid, dashed and dotted lines, in addition to no interaction, represented by thin grey lines. In the \textsc{nri} model, the possible combinations of interactions require eight ($=2^3$) edge-types.}
\label{fig:factorisation}%
\end{figure*}

\subsection{Factorised Neural Relational Inference}
\label{S:fNRI}

Here we introduce our reformulation of the \textsc{nri} model which we will refer to as the factorised neural relational inference (f\textsc{nri}) model. In this model the \textsc{nri}'s single latent interaction graph with $K$ edge-types is \textit{factorised} into an $n$-layer multiplex graph (see figure \ref{fig:factorisation}), where the $a$-th layer-graph has $K_a$ edge-types.



The $K$-dimensional edge-embedding vector $\mathbf{h}^{2}_{(i,j)}$ returned by the \textsc{nri} encoder (as in equation \ref{eq:intro_NRI_encoder}) is first segmented
\begin{align}
\mathbf{h}^{2}_{(i,j)} = \big[\mathbf{h}^{2,1}_{(i,j)},\dots,\mathbf{h}^{2,n}_{(i,j)}\big]
\end{align}
where segment $\mathbf{h}^{2,a}_{(i,j)}$ is a $K_a$-dimensional vector and $K = \sum^n_{a=1} K_a$ is the total number of edge types. The posterior distribution for each layer-graph is then formed as
\begin{align}
q_{\theta}(\mathbf{z}^a_{ij}|\mathbf{x}) = \text{softmax}(\mathbf{h}^{2,a}_{(i,j)})
\end{align}
where $\mathbf{z}^a_{ij}$ denotes the one-hot edge-type vector between objects $i$ and $j$ in the $a$-th layer-graph. As in the \textsc{nri}, during training the vectors are sampled from a `continuous relaxation' of their respective posterior distributions using the concrete distribution \cite{Concrete}
\begin{align}
\mathbf{z}^a_{ij} = \text{softmax}\big( (\mathbf{h}^{2,a}_{(i,j)} + \mathbf{g}) / {\tau} \big). \nonumber
\end{align}
where $\mathbf{g} \in \mathbb{R}^{K_a}$ is a vector of i.i.d samples drawn from a $\text{Gumbel}(0,1)$ distribution and $\tau$ is the `softmax temperature.' Concatenating these vectors forms the combined edge-type vector of the multiplex interaction graph
\begin{align}
\mathbf{z}_{ij} = [ \mathbf{z}^1_{ij}, ..., \mathbf{z}^n_{ij} ].
\end{align} 
These $\mathbf{z}_{ij}$ are no longer one-hot vectors, but rather multi-categoric with $\sum_k z_{ij,k} = n$, and are supplied to the \textsc{nri} decoder as described in \ref{S:nri_overview}. In alignment with the \textsc{nri} model, the latent graphs are not forced to be undirected ($\mathbf{z}_{ij}$ may not necessarily equal $\mathbf{z}_{ji}$), and if desired, the first edge-type of each layer-graph can be hard-coded as the non-edge. The \textsc{kl}-divergence term in the \textsc{elbo} is the sum of \textsc{kl}-divergences over the layer-graphs.
\begin{align}
D_{\textsc{kl}}[ q_{\theta}(\mathbf{z}|\mathbf{x})||p(\mathbf{z}) ] = \sum_{a=1}^n D_{\textsc{kl}}[ q_{\theta}(\mathbf{z}^a|\mathbf{x})||p(\mathbf{z}^a) ]
\end{align}



\subsubsection{Motivations}

We now expand on the motivations given earlier in light of the model specifications. As the \textsc{nri} uses a one-hot latent encoding, in multi-interaction systems single edge-types must exist to represent any possible combination of interactions (e.g. \textit{spring}$+$\textit{charge}). In contrast, the f\textsc{nri} edge-types need only encode for one interaction-type, with combinations arising naturally from the multiplex structure. The edge-type decoder networks $\tilde{f}_e^k$ in equation (\ref{eq:intro_NRI_decoder1}) therefore only need to decode the dynamics of a single type of interaction. We theorise this compartmentalisation of the interactions will improve training in complex systems, especially given that each of the networks $\tilde{f}_e^k$ will effectively have a larger training set in our formulation. This is because the $\tilde{f}_e^k$ are used by the decoder in \textit{every} instance that its corresponding interaction is present, rather than when a specific combination of interactions is present, as in the \textsc{nri}. Or in other words, because the density of the latent representation is exponentially greater in the f\textsc{nri} model. This increase in latent information density also means that factorised model decoders have notably fewer parameters. 

In addition, the f\textsc{nri} model has the capacity to be explicitly fractionally correct about an edge-type. If the encoder correctly predicts one underlying interaction type, but the other incorrectly, in the \textsc{nri} model the corresponding $\mathbf{z}_{ij}$ is plainly `incorrect'. However in the f\textsc{nri} model, the corresponding $\mathbf{z}_{ij}$ will be half-right and treated accordingly, in theory allowing for better directed feedback.

Compartmentalising interactions in the f\textsc{nri} model will also be useful when attempting to understand the meanings of edge-types in systems where the underlying interactions are unknown. An issue that could be raised with the f\textsc{nri} model is that in such contexts, due to having $ \lbrace K_1, ..., K_n \rbrace $ edge-types rather than just $K$, the dimensionality of the hyperparameter space has been increased. However, picking $K_a = 2$ for all $a$ allows for the same dimensionality while retaining all functionality, where interactions with more than two discrete edge-types (e.g. colour-charge) are encoded over multiple layer-graphs.

\subsection{Sigmoid Factorisation}

We also investigate a drastic simplification of the f\textsc{nri} model, where each layer-graph effectively only contains a single edge-type and probabilistic sampling is removed completely. In this sf\nri model, rather than using the edge-embedding vectors $\mathbf{h}^2_{(i,j)}$ returned by the encoder to form posterior distributions, they are directly transformed into $K$-dimensional edge-type vectors by a sigmoid function $\mathbf{z}_{ij} = \sigma(\mathbf{h}^2_{(i,j)})$. These $\mathbf{z}_{ij}$ are then decoded using the same decoder described in section \ref{S:nri_overview}. In this model there are $K$ layer-graphs, each of which contains a \textit{single} edge-type, in addition to an explicit non-edge.

As the sampling aspect of the model is removed, the elements of the edge-type vectors $z_{ij,k}$ are no longer strictly binary elements of $\lbrace 0,1 \rbrace$, but rather are elements of $[0,1]$. Furthermore, it is no longer possible to define a \textsc{kl}-divergence so the loss function is just the reconstruction error -- a rescaling of the mean squared error between the predicted and ground-truth trajectories.

The motivations here are much the same as for the f\textsc{nri}; allow each element of the edge-type vector $\mathbf{z}_{ij}$ to represent a separate interaction edge that can be observed in combination. Additionally, the non-interaction edge becomes a more fundamental part of the model. When there are no interactions between a pair of particles, the ground truth edge-vector will, in theory, be all zeros, $\mathbf{z}_{ij}=\mathbf{0}$. This follows as if a particle has no interactions, then the elements of the vector $\sum_{i \neq j} \tilde{\mathbf{h}}^t_{(i,j)}$ in equation (\ref{eq:intro_NRI_decoder2}) will all be zero, and the only non-zero entries to the neural network $\tilde{f}_v$ will be the current state of the particle $\mathbf{x}^t_j$. This means that the non-interaction graph (where there are no interactions between particles) is made explicit by the very architecture of the model, as $\mathbf{z}$ will contain only zeros and therefore each particle's predicted future state $\boldsymbol{\mu}_j^{t+1}$ can \textit{only} depend on its current state $\mathbf{x}^t_j$.

\begin{table*}[h]
\small
    \centering
        \caption{\footnotesize Accuracy (\%) in recovering the ground truth interaction graph. Higher is better.}
        \label{tab:edge_acc}
        \sisetup{%
            table-align-uncertainty=true,
            separate-uncertainty=true,
            detect-weight=true,
            detect-inline-weight=math
        }
        \begin{tabular}{l|*{3}{S[table-format=2.1]@{\,\(\tinypm \)\,}S[table-format=1.1]}|*{4}{S[table-format=2.1]@{\,\(\tinypm \)\,}S[table-format=1.1]}}
            \toprule
            & \multicolumn{6}{c|}{\ubold I-Springs+Charges} & \multicolumn{8}{c}{\ubold I-Springs+Charges+F-springs} \\
            \midrule
            Accuracy & \multicolumn{2}{c}{\ubold Combined} & \multicolumn{2}{c}{I-Springs} & \multicolumn{2}{c|}{Charges} & \multicolumn{2}{c}{\ubold Combined} & \multicolumn{2}{c}{I-Springs} & \multicolumn{2}{c}{Charges} & \multicolumn{2}{c}{F-Springs} \\
            \midrule
            \vspace{1mm}
            Random & \multicolumn{2}{c}{25.0} & \multicolumn{2}{c}{50.0} & \multicolumn{2}{c|}{50.0} & \multicolumn{2}{c}{12.5} & \multicolumn{2}{c}{50.0} & \multicolumn{2}{c}{50.0} & \multicolumn{2}{c}{50.0} \\
            \textsc{nri} \ \ (learned) & 89.1 & \scriptsize 0.4 & 97.9 & \scriptsize 0.0 & 91.0 & \scriptsize 0.4 & 57.9 & \scriptsize 6.1 & 88.5 & \scriptsize 0.9 & 87.3 & \scriptsize 6.2 & 70.7 & \scriptsize 2.3 \\
            f\textsc{nri} \ (learned) & \ubold 94.0 & \scriptsize \ubold 1.4 & 98.0 & \scriptsize 0.1 & 95.8 & \scriptsize 1.3 & \ubold 63.3 & \scriptsize \ubold 6.5 & 86.9 & \scriptsize 2.7 & 97.7 & \scriptsize 0.7 & 69.2 & \scriptsize 5.5 \\
            \vspace{1mm}
            sf\textsc{nri} (learned) & 88.8 & \scriptsize 0.8 & 97.6 & \scriptsize 0.1 & 91.1 & \scriptsize 0.8 & 45.1 & \scriptsize 5.1 & 90.0 & \scriptsize 2.3 & 98.2 & \scriptsize 0.8 & 52.4 & \scriptsize 2.7 \\
            \textsc{nri} \ \ (supervised) & \ubold 98.3 & \scriptsize \ubold 0.0 & 98.6 & \scriptsize 0.0 & 99.7 & \scriptsize 0.0 & 80.9 & \scriptsize 0.7 & 92.4 & \scriptsize 0.3 & 99.0 & \scriptsize 0.1 & 84.4 & \scriptsize 0.4 \\
            f\textsc{nri}  \ (supervised) & \ubold 98.3 & \scriptsize \ubold 0.0 & 98.8 & \scriptsize 0.4 & 99.4 & \scriptsize 0.4 & \ubold 81.8 & \scriptsize \ubold 0.1 & 93.3 & \scriptsize 0.1 & 99.3 & \scriptsize 0.0 & 85.8 & \scriptsize 0.1 \\
            sf\textsc{nri} (supervised) & 98.0 & \scriptsize 0.0 & 98.3 & \scriptsize 0.0 & 99.6 & \scriptsize 0.0 & 81.0 & \scriptsize 0.3 & 92.9 & \scriptsize 0.1 & 99.2 & \scriptsize 0.0 & 85.2 & \scriptsize 0.2 \\
            \bottomrule
        \end{tabular}
    
\end{table*}

\begin{table*}
\small
    \centering
    \caption{\footnotesize Mean squared error (\textsc{mse}) $/ \ 10^{-5}$ in trajectory prediction. Lower is better.}
    \label{tab:reconstruction}
        \sisetup{%
            table-align-uncertainty=true,
            separate-uncertainty=true,
            detect-weight=true,
            detect-inline-weight=math
        }
        \begin{tabular}{l|*{2}{S[table-format=1.2]@{\,\( \tinypm \)\,}S[table-format=1.2]}S[table-format=2.2]@{\,\( \tinypm \)\,}S[table-format=1.2]|*{2}{S[table-format=1.2]@{\,\(\tinypm \)\,}S[table-format=1.2]}S[table-format=2.2]@{\,\( \tinypm \)\,}S[table-format=1.2]}
      \toprule
      & \multicolumn{6}{c|}{\textbf{ I-Springs+Charges}} & \multicolumn{6}{c}{\textbf{ I-Springs+Charges+F-Springs}}\\
      \midrule
    Predictions Steps & \multicolumn{2}{c}{1} & \multicolumn{2}{c}{10} & \multicolumn{2}{c|}{20} & \multicolumn{2}{c}{1} & \multicolumn{2}{c}{10} & \multicolumn{2}{c}{20} \\
        \midrule
        \vspace{1mm}
        Static & \multicolumn{2}{c}{19.4} & \multicolumn{2}{c}{283} & \multicolumn{2}{c|}{783} & \multicolumn{2}{c}{12.8} & \multicolumn{2}{c}{274} & \multicolumn{2}{c}{782} \\
        \textsc{nri} \ \ (learned) & 0.88 & \scriptsize 0.06 & 4.05 & \scriptsize 0.22 & 11.5 & \scriptsize 0.5 & 0.95 & \scriptsize 0.05 & 8.67 & \scriptsize 0.45 & 29.1 & \scriptsize 1.4 \\
        f\textsc{nri} \ (learned) & \ubold 0.80 & \scriptsize \ubold 0.04 & 3.54 & \scriptsize 0.09 & 9.93 & \scriptsize 0.29 & 0.81 & \scriptsize 0.05 & 7.78 & \scriptsize 0.20 & 26.8 & \scriptsize 0.8 \\
        \vspace{1mm}
        sf\textsc{nri} (learned) &  1.03 & \scriptsize 0.09 & \ubold 3.32 & \scriptsize \ubold 0.23 & \ubold 9.68 & \scriptsize \ubold 0.74 & \ubold 0.77 & \scriptsize \ubold 0.03 & \ubold 5.69 & \scriptsize \ubold 0.21 & \ubold 19.3  & \scriptsize \ubold 0.8 \\
        \textsc{nri} \ \ (true graph) & 0.85 & \scriptsize 0.04 & 1.59 & \scriptsize 0.26 & 3.20 & \scriptsize 0.15 & 0.75 & \scriptsize 0.02 & 1.55 & \scriptsize 0.07 & 3.43 & \scriptsize 0.21 \\
        f\textsc{nri}  \ (true graph) & \ubold 0.70 & \scriptsize \ubold 0.03 & \ubold 1.30 & \scriptsize \ubold 0.06 & \ubold 2.52 & \scriptsize \ubold 0.11 & \ubold 0.51 & \scriptsize \ubold 0.05 & 0.97 & \scriptsize 0.08 & 2.44 & \scriptsize 0.28 \\
        sf\textsc{nri} (true graph) & 0.86 & \scriptsize 0.09 & 1.32 & \scriptsize 0.06 & 2.77 & \scriptsize 0.07 & 0.56 & \scriptsize 0.04 & \ubold 0.89 & \scriptsize \ubold 0.06 & \ubold 2.28 & \scriptsize \ubold 0.15 \\
    \bottomrule
    \end{tabular}
\end{table*}

\section{Experiments}

To make our comparison with the original \textsc{nri} model as convincing as possible, unless otherwise stated we use the exact same hyperparameters as detailed in the original paper \cite{Kipf2018NeuralSystemsb}, full details of which can be found in the supplementary material. The only change we make to the training routine is discussed in section \ref{S:compression}.\footnote{Our implementation is available in full at \url{https://github.com/ekwebb/fNRI}.}

We experiment with simulated systems of 5 interacting particles in a finite 2D box. In these systems particles are `randomly connected' by different physical interactions. We consider three different types of physical interaction: \textit{ideal springs} (I-springs) where particles are randomly connected by Hookean springs of zero length, \textit{finite springs} (F-springs) where particles are randomly connected by Hookean springs of a fixed finite length, and \textit{charges} where particles are randomly selected to be either positively charged or neutral, and charged particles interact via Coulomb's law. 

\subsection{Compression Models}
\label{S:compression}

A problem we encounter with the \textsc{nri} training routine is that when attempting to learn more complex interaction graphs, the encoder can instead learn to use the latent space to store a compressed version of the input trajectories. It appears that this can occur to a varying degree, however the problem worsens as the size, and thus the expressiveness, of the latent space increases. These models are easily identified during testing as they are non-predictive, meaning they can only reconstruct the trajectories the encoder received as input.

In order to avoid these compression models, we modify the training routine such that the encoder receives the first half of the particle trajectories, and the decoder predicts the second half of the particle trajectories. For interacting systems with static interaction graphs, this change is reasonable, and has a number of distinct advantages. Firstly, compression solutions are avoided as the models are now trained to predict \textit{unobserved} trajectories, only. As such, training becomes significantly more reliable and far less dependent on the model initialisation. Secondly, the difference in the reconstruction loss between the training and validation sets is reduced, and we observe a reduction in overfitting. Making this change means the network is formally no longer acting as an auto-encoder, as the decoder network does not learn by \textit{reconstructing} the encoder input $\mathbf{x}$, but rather by generating a time-evolution of $\mathbf{x}$, which is then compared to the ground-truth time-evolution. We use this modification when training \textit{all} the models presented here. Without it, training is simply not reliable enough, with edge-accuracies often failing to rise above the random level.

\section{Results}

The edge and trajectory prediction results are summarised in tables \ref{tab:edge_acc} and \ref{tab:reconstruction} respectively, where each result is the average over 5 runs with the standard error given. In all cases the factorised \textsc{nri} models match or outperform the original.

For both edge and trajectory prediction, we compare the unsupervised \textit{learned} models to the supervised `gold standards.' For edge prediction the \textit{supervised} encoders are trained in isolation on the ground-truth interaction graphs, and for trajectory prediction the \textit{true graph} decoders are trained in isolation with the ground-truth interaction graphs their inputs. The static decoder simply returns the state vector it receives as input. For edge prediction, accuracies are decomposed into the prediction accuracy for each interaction type. The combined accuracy is calculated such that it only receives a contribution when the predicted edges between a pair of nodes are correct for all interaction types. 

\section*{Acknowledgements}
We would like to thank Thomas Kipf, Ethan Fetaya, Kuan-Chieh Wang, Max Welling \& Richard Zemel for making the codebase for the \textsc{nri} model \cite{Kipf2018NeuralSystemsb} publicly available. This work was made possible by their commitment to open research practices. We would also like to thank the developers of PyTorch \cite{paszke2017automatic}.

\bibliography{example_paper,references,additionalreferences}
\bibliographystyle{icml2019}

\newpage
\twocolumn[
    \begin{@twocolumnfalse}
    \icmltitle{Factorised Neural Relational Inference for Multi-Interaction Systems: \\
    Supplementary Material}
    \end{@twocolumnfalse}
]

\section*{Overview}
These supplementary materials are provided to support the workshop paper `Factorised Neural Relational Inference for Multi-Interaction Systems' published at the Learning and Reasoning with Graph-Structured Data workshop at ICML 2019.

The materials include an extended description of the \textsc{nri} model in section \ref{nri_section}, details of the physics simulations and experimental procedures in sections \ref{simulations_section} and \ref{S:experimental_section}, and a note on calculating edge accuracy in unsupervised systems is added in section \ref{edge_section}.

\section{Neural Relational Inference}
\label{nri_section}
Here we describe the \textsc{nri} model as presented by Kipf et al. \yrcite{Kipf2018NeuralSystems} along with our own schematic and comments. Here we provide an extended description of the \textsc{nri} model; adopting the formalism and nomenclature of Kipf et al. \yrcite{Kipf2018NeuralSystems} throughout. The \nri model takes the generalised form of a variational auto-encoder (\textsc{vae}), where the encoding network infers a latent interaction graph for the system, and the decoding network predicts the future dynamics of the system using this interaction graph. This graph is described by a set of edge-types $\mathbf{z}$ (the latent variables of the \textsc{vae}) which tell the decoding network about the types of interactions between each pair of particles.

\begin{figure*}[ht!]
\centering
\includegraphics[width=16cm ,trim=0cm 0.8cm 0cm 0.5cm]{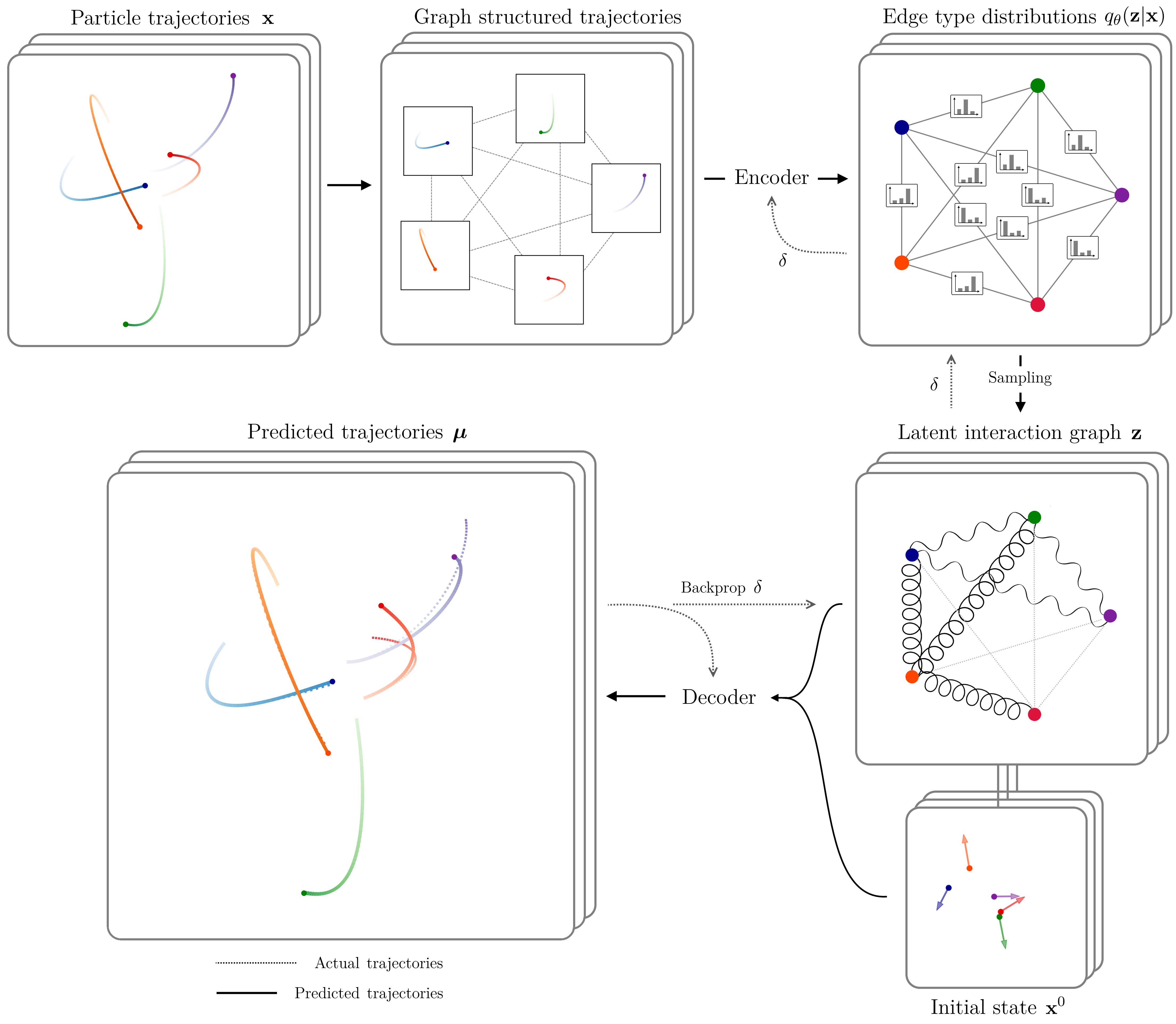}
\caption{\footnotesize Schematic of the batch-wise \textsc{nri} training procedure, where the dashed arrows with $\delta\,$s indicate backpropagation. The system in the schematic has 5 interacting particles and three inferred edge-types. In the latent interaction graph $\mathbf{z}$ these three different edge-types are represented by sinusoidal lines, curly lines and thin grey lines.}
\label{fig:NRI_schematic}
\end{figure*}

The \textsc{nri} model differs from the standard \textsc{vae} implementation in a number ways. Most notably, it does not use a continuous isotropic multivariate Gaussian distribution as its prior. Rather, its prior distribution is \textit{discrete}; and the encoder returns a probability vector for the edge-type between each pair of particles. The edge-types $\mathbf{z}$ in the latent interaction graph are then sampled from these probability vectors (see figure \ref{fig:NRI_schematic}). 

In order for the \textsc{nri} model to be successful in predicting the future dynamics of a system, the underlying interactions of the system must be \textit{discrete}. In the context of physics, this means that the interactions must be discrete in both form and strength. For example, if we have a box containing a collection of interacting charged particles, the \textsc{nri} model has the potential to successfully model the dynamics of this system provided the strengths of the charges are picked from some finite set, rather than being picked from a continuum. The reason for this is that the number of edge-types $K$ in the latent interaction graph, is a hyperparameter of the model and represents the number of distinct `interaction types' the model will be able to encode for. If the strength of the charges are drawn from a continuum, although interactions will be discrete in form (with all the forces between particles being proportional to the inverse square of their separation), an interaction graph cannot be drawn for the system using a discrete set of edge-types.\footnote{At least one cannot be drawn using a set of edge-types that is smaller than the total number of edges in the interaction graph.}

In the latent interaction graph, the edge-type between objects $i$ and $j$ is encoded for using a one-hot vector of length $K$, denoted $\mathbf{z}_{ij}$. This means each edge in the interaction graph is one of $K$ \textit{discrete} edge-types, formalised as $\sum_{k=1}^K z_{ij,k} = 1$, where $z_{ij,k} \in \lbrace 0, 1 \rbrace$ denotes the $k$-th element of the vector $\mathbf{z}_{ij}$.

\subsection{Message Passing Operation}
\label{S:NRI_GNN}

The encoding and decoding networks in the \textsc{nri} model are described as graph neural networks (\textsc{gnn}s). These are a broad class of artificial neural networks which operate on graph structured data and are defined by their use of the `message passing' operation introduced by Gilmer et al. (\citeyear{pmlr-v70-gilmer17a}). For a graph $\mathcal{G} = (\mathcal{V},\mathcal{E})$ with vertices $v \in \mathcal{V}$ and edges $e = (v, v') \in \mathcal{E}$, where vertex $v_i$ has features $\mathbf{x}_i$ and edge $e_{(i,j)}$ has features $\mathbf{x}_{(i,j)}$, a single node-to-node message passing operation is defined as
\begin{align}
v \rightarrow e \colon \ \ \ \mathbf{h}^l_{(i,j)} &= f^l_e\big( \big[ \mathbf{h}^l_i, \mathbf{h}^l_j, \mathbf{x}_{(i,j)} \big] \big) \\
e \rightarrow v \colon \ \ \ \ \mathbf{h}^{l+1}_j &= f^l_v\big( \big[ \textstyle\sum\nolimits_{i \in \mathcal{N}_j} \mathbf{h}^l_{(i,j)}, \mathbf{x}_j \big] \big)
\end{align}
where $\mathbf{h}^{l}_j$ is the embedding of the features of vertex $v_i$ in layer $l$ of the \textsc{gnn} and $\mathbf{h}^l_{(i,j)}$ is the embedding of the features of edge $e_{(i,j)}$ in layer $l$ of the \textsc{gnn}. These edge feature embeddings are sometimes referred to as a `messages'. $\mathcal{N}_j$ denotes the set of indices of vertices which are connected to vertex $v_j$ by an incoming edge, and $[\cdot,\cdot]$ denotes concatenation of vectors. The functions $f_v$ and $f_e$ are node- and edge-specific neural networks respectively, for example small multi-layer perceptrons (\textsc{mlp}s). We note that the message passing operation operates on the edge and node \textit{features}, and does not alter the shape of the graph.

\subsection{Encoder}
\label{S:NRI_encoder}

The input of the encoder consists of the trajectories of $N$ objects. We denote the feature vector of object $i$ at time $t$ by $\mathbf{x}^t_i$; in our work this vector contains the location and velocity of the particle. We denote the set of all $N$ objects at time $t$ by $\mathbf{x}^t = \lbrace \mathbf{x}^t_1, ..., \mathbf{x}^t_N \rbrace$, the trajectory of object $i$ by $\mathbf{x}_i = ( \mathbf{x}^1_i, ..., \mathbf{x}^T_i )$ and the set of all trajectories by $\mathbf{x} = ( \mathbf{x}^1, ..., \mathbf{x}^T )$, where $T$ is the total number of time steps.

The trajectory of each particle enters the encoder as the features of a node in a fully-connected graph of $N$ nodes, where each node represents one of the interacting objects. Using the message passing operations defined in section \ref{S:NRI_GNN}, the action of the encoding network on this graph can be defined as follows:
\begin{alignat}{2}
\label{eq:NRI_encoder1}
& & \mathbf{h}^1_i &= f_{\text{emb}}(\mathbf{x}_i) \\
\label{eq:NRI_encoder2}
v \rightarrow e \colon \ \ & & \mathbf{h}^1_{(i,j)} &= f^1_e\big( \big[ \mathbf{h}^1_i, \mathbf{h}^1_j \big] \big) \\
\label{eq:NRI_encoder3}
e \rightarrow v \colon \ \ & & \mathbf{h}^2_j &= f^1_v\big( \textstyle\sum\nolimits_{i \neq j} \mathbf{h}^1_{(i,j)} \big) \\
\label{eq:NRI_encoder4}
v \rightarrow e \colon \ \ & & \mathbf{h}^2_{(i,j)} &= f^2_e\big( \big[ \mathbf{h}^2_i, \mathbf{h}^2_j \big] \big)
\end{alignat}
The edge-type posterior distributions are then taken as $q_{\theta}(\mathbf{z}_{ij}|\mathbf{x}) = \text{softmax}(\mathbf{h}^2_{(i,j)})$, where $\mathbf{h}^2_{(i,j)} \in \mathbb{R}^K$ and $\theta$ summarizes the parameters of the neural networks in equations (\ref{eq:NRI_encoder1})-(\ref{eq:NRI_encoder4}). By studying equations (\ref{eq:NRI_encoder1})-(\ref{eq:NRI_encoder4}), it can be noted that as the input graph is fully connected, the node embeddings $\mathbf{h}^2_j$ and subsequent edge embeddings $\mathbf{h}^2_{(i,j)}$ are influenced by the trajectories of all the particles in the system.

The neural networks $f_{\text{emb}}$, $f^1_e$ and $f^1_v$ are 2-layer \textsc{mlp}s with hidden and output dimension 256, batch normalization, and \textsc{elu} activations. The last neural network $f^2_e$ has these same properties with the addition of an extra dense layer of output dimension $K$.

\subsection{Sampling}
\label{S:NRI_sampling}

A softmax function is used to transform the edge feature embedding $\mathbf{h}^2_{(i,j)}$ into a posterior distribution $q_{\theta}(\mathbf{z}_{ij}|\mathbf{x})$. However, sampling directly from this distribution is not a differentiable process. To circumvent this problem, the \textsc{nri} model uses a `continuous relaxation' of the discrete posterior distribution in the form of the concrete distribution \cite{Concrete}, which \textit{reparametrises} the sampling using the Gumbel distribution. This means rather than sampling the edge-type vectors $\mathbf{z}_{ij}$ directly from posterior as
\begin{align}
\mathbf{z}_{ij} \sim q_{\theta}(\mathbf{z}_{ij}|\mathbf{x}) = \text{softmax}(\mathbf{h}^2_{(i,j)})
\end{align}
The edge-type vectors are sampled using
\begin{align}
\label{eq:concrete}
\mathbf{z}_{ij} = \text{softmax}\big( (\mathbf{h}^2_{(i,j)} + \mathbf{g}) / {\tau} \big)
\end{align}
where $\mathbf{g} \in \mathbb{R}^K$ is a vector of independent samples drawn from a Gumbel(0,1) distribution and $\tau$ is the softmax temperature. This is a \textit{continuous relaxation} of the discrete posterior distribution $q_{\theta}(\mathbf{z}_{ij}|\mathbf{x})$ as the edge-type vectors $\mathbf{z}_{ij}$ returned by equation (\ref{eq:concrete}) are not one-hot, but rather smoothly converge to one-hot vectors sampled from $q_{\theta}(\mathbf{z}_{ij}|\mathbf{x})$ in the limit $\tau \rightarrow 0$.

\subsection{Decoder}
\label{S:NRI_decoder}

The task of the decoder is to predict the future dynamics of the system using the latent interaction graph and the past dynamics. Formally, this means calculating the likelihood $p_{\phi}(\mathbf{x}^{t+1}|\mathbf{x}^t; ... ; \mathbf{x}^1; \mathbf{z})$. In our work we only consider systems where the dynamics are Markovian, meaning the dependence in the likelihood reduces to $p_{\phi}(\mathbf{x}^{t+1}|\mathbf{x}^t; \mathbf{z})$.

In the \textsc{nri} model, each edge-type has a separate neural network in the edge-to-vertex message passing operation. The message passing section of the Markovian decoder is formalised as:
\begin{align}
\label{eq:NRI_decoder1}
v \rightarrow e: \tilde{\mathbf{h}}^t_{(i,j)} &= \sum_{k=1}^K z_{ij,k} \tilde{f}^k_e \big( [\mathbf{x}_i^t,\mathbf{x}_j^t] \big) \\
e \rightarrow v: \boldsymbol{\mu}_j^{t+1} &= \mathbf{x}_j^t + \tilde{f}_v \big( \big[ \textstyle\sum\nolimits_{i \neq j} \tilde{\mathbf{h}}^t_{(i,j)}, \mathbf{x}_j^t \big] \big)
\label{eq:NRI_decoder2}
\end{align}
The future state of each object is then sampled from an isotropic Gaussian distribution with a mean vector $\boldsymbol{\mu}_j^{t+1}$ and a fixed (user-defined) variance $\sigma^2$:
\begin{align}
p_{\phi}(\mathbf{x}_j^{t+1}|\mathbf{x}^t,\mathbf{z}) = \mathcal{N}( \boldsymbol{\mu}_j^{t+1}, \sigma^2 \mathbf{I} )
\end{align}
Note that in equation (\ref{eq:NRI_decoder1}), when the edge-type vector $\mathbf{z}_{ij}$ is one-hot, $\tilde{\mathbf{h}}^t_{(i,j)}$ only receives a contribution from the neural network representing the `hot' edge-type, but for continuous relaxations, the message is a weighted sum. We note that the first edge-type can be `hard-coded' to be the non-edge, representing no interaction between particles, by modifying the sum in equation (\ref{eq:NRI_decoder1}) to start at $k = 2$.

When the dynamics of the system are not Markovian, a recurrent neural network can be used in the decoder to use the full history of the particle in predicting its future dynamics.

\subsection{Training}
\label{S:NRI_training}

The \textsc{nri} model takes the form of a variational auto-encoder and it is therefore trained to maximise the evidence lower bound
\begin{align}
\mathcal{L} =\ \ &\mathbb{E}_{q_{\theta}(\mathbf{z}|\mathbf{x})}[ \log p_{\phi}(\mathbf{x}|\mathbf{z}) ] - D_{\textsc{kl}}[ q_{\theta}(\mathbf{z}|\mathbf{x})||p(\mathbf{z}) ]
\end{align}
where the likelihood $p_{\phi}(\mathbf{x}|\mathbf{z})$ can be expanded as $p_{\phi}(\mathbf{x}|\mathbf{z}) = \prod_{t=1}^Tp_{\phi}(\mathbf{x}^{t+1}|\mathbf{x}^t;\mathbf{z})$, and the prior $p(\mathbf{z}) = \prod_{i \neq j} p(\mathbf{z}_{ij})$ is generally a factorised uniform distribution over edge-types. For a uniform prior, $p(z_{ij,k}) = 1/K$, the overall \textsc{kl}-divergence in the \textsc{elbo} function is given by
\begin{align}
D_{KL}[ q_{\theta}(\mathbf{z}|\mathbf{x})||p(\mathbf{z}) ] &= \sum_{i \neq j} \Big[ -H \big[ q_{\theta}(\mathbf{z}_{ij}|\mathbf{x}) \big] + \log K \Big]
\end{align}
where $H \big[ q_{\theta}(\mathbf{z}_{ij}|\mathbf{x}) \big]$ is the entropy of the posterior distribution $q_{\theta}(\mathbf{z}_{ij}|\mathbf{x})$. The reconstruction error in the ELBO is estimated by
\begin{align}
\mathbb{E}_{q_{\theta}(\mathbf{z}|\mathbf{x})}[ \log p_{\phi}(\mathbf{x}|\mathbf{z}) ] = -\sum_j \sum_{t=2}^T \frac{||\mathbf{x}_j^t-\boldsymbol{\mu}_j^t||^2}{2\sigma^2} + \text{const}
\end{align}
This reconstruction error only depends on single time step predictions. However, the interactions between objects often only have a small effect on the short term dynamics. This means the decoder could quite easily learn to ignore the latent interaction graph, whilst achieving only a marginally worse reconstruction error. In order to avoid these `degenerate' decoders, the \textsc{nri} model predicts the dynamics multiple time-steps in to future. Denoting the decoder as $\boldsymbol{\mu}_j^{t+1} = f_{\text{dec}}(\mathbf{x}^t_j)$, the \textsc{nri} model implements this by replacing the actual system state $\mathbf{x}^t$ with the previous predicted mean state $\boldsymbol{\mu}_j^{t}$ for $M$ time-steps. Doing this means that any errors in the reconstruction accumulate over $M$ steps, which makes correctly predicting the latent interaction graph essential for maximising the \textsc{elbo}. This procedure can be formalised as
\begin{align*}
\boldsymbol{\mu}_j^{2} &= f_{\text{dec}}(\mathbf{x}^1_j) \\
\boldsymbol{\mu}_j^{t+1} &= f_{\text{dec}}(\boldsymbol{\mu}^t_j) \qquad \quad t = 2, ..., M \\
\boldsymbol{\mu}_j^{M+2} &= f_{\text{dec}}(\mathbf{x}^{M+1}_j) \\
\boldsymbol{\mu}_j^{t+1} &= f_{\text{dec}}(\boldsymbol{\mu}^t_j) \qquad \quad t = M + 2, ..., 2M \\
& \ ...
\end{align*} 
If we have some prior knowledge of the system, this can be included in the form a non-uniform prior. For example, when the first edge-type is hard-coded to be the non-edge, a non-uniform prior with a higher probability on the non-edge could be used to encourage sparser graphs.

\section{Simulations}
\label{simulations_section}
In accordance with the work by Kipf et al. \yrcite{Kipf2018NeuralSystems}, we simulate $N = 5$ point mass particles in a finite 2D box, where collisions with the box wall are elastic  and there are no external forces. The initial locations of the particles are sampled from a Gaussian distribution $\mathcal{N}(0,0.5)$, and the initial velocity of each particle is a random vector with norm 0.5. We consider 3 different types of particle interactions in this investigation:
\paragraph{Ideal spring interactions} where particles connected by an ideal spring are acted on by forces given by Hooke's law 
\begin{align}
\mathbf{F}_{ij} = -k_I( \mathbf{r}_i - \mathbf{r}_j)
\end{align} 
where $\mathbf{F}_{ij}$ is the force applied to particle $i$ by particle $j$, $k_I$ is the spring constant, and $\mathbf{r}_i$ is the 2D location vector of particle $i$. These are `ideal springs' (I-springs) because they have zero length and are therefore only attractive.
\paragraph{Finite spring interactions} where particles connected by a finite length spring are acted on by forces given by a modified Hooke's law 
\begin{align}
\mathbf{F}_{ij} = -k_F\Big( \mathbf{r}_i - \mathbf{r}_j - l \cdot \frac{\mathbf{r}_i - \mathbf{r}_j}{|\mathbf{r}_i - \mathbf{r}_j|} \Big)
\end{align} 
where $k_F$ is the spring constant and $l$ is the spring length. The forces these finite length springs (F-springs) generate between particles can be attractive or repulsive.
\paragraph{Charge interactions} where charged particles are acted on by forces given by Coulomb's Law
\begin{align}
\mathbf{F}_{ij} = q_i q_j C \cdot \frac{\mathbf{r}_i-\mathbf{r}_j}{|\mathbf{r}_i-\mathbf{r}_j|^3}
\end{align} 
where $C$ is a positive constant and $q_i$ is the charge of particle $i$. Due to the simulation instabilities that arise when divergent forces are present, in our investigation we only consider repulsive charge interactions where $q_i \in \lbrace 0, +1 \rbrace$.

We combine these three interaction types in two different ways to form two types of simulated system. In the \textsc{i+c} system, ideal spring and charge interactions are randomly added between particles, and in the \textsc{i+c+f} system, ideal spring, charge and finite spring interactions are randomly added between particles. The procedure for this random interaction assignment is described below.

\begin{figure}[t!]
\centering
\includegraphics[width=8.3cm ,trim=0.4cm 1.7cm 0cm 0.4cm]{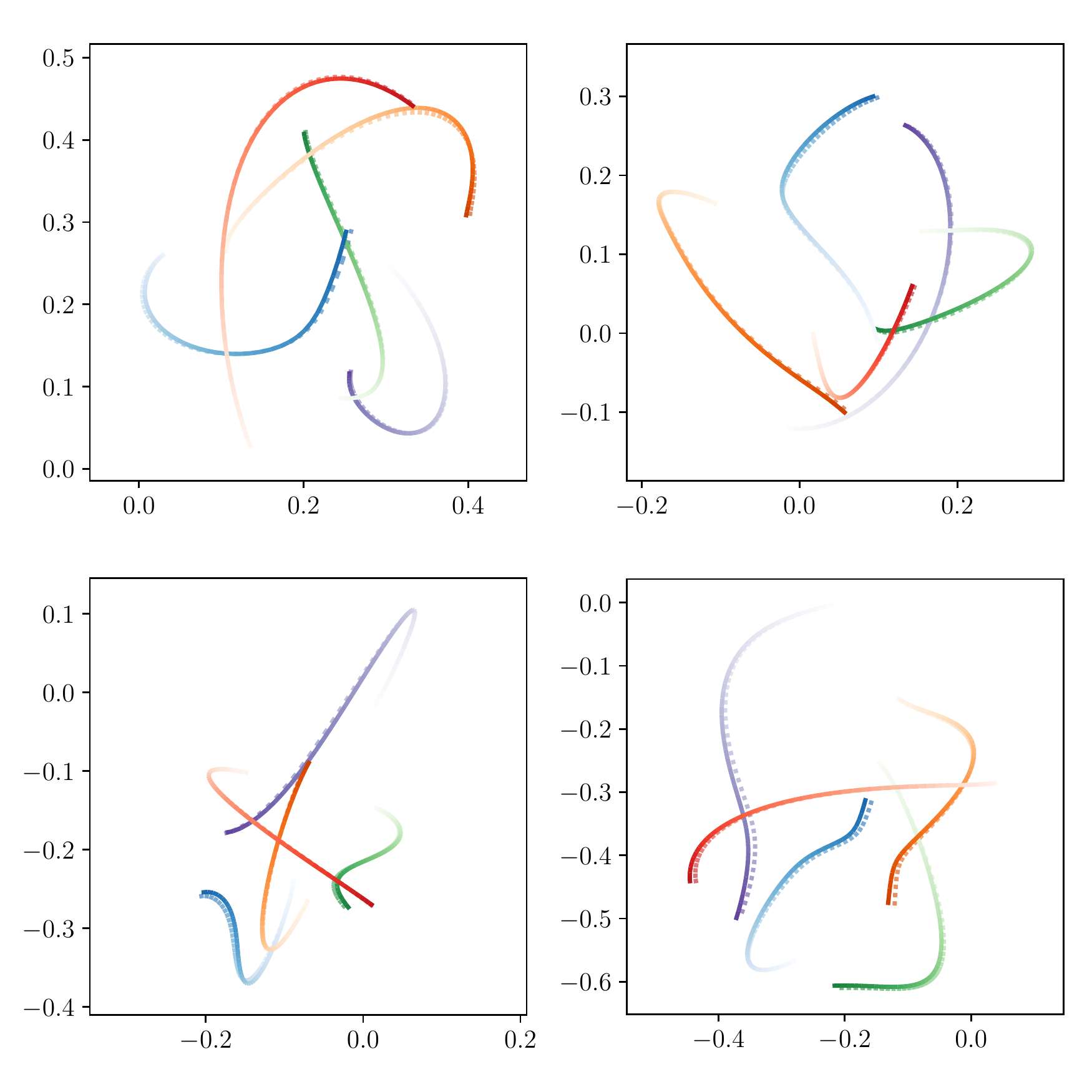}
\caption{\footnotesize Trajectories of four sets of 5 interacting particles in the \textsc{i+c} system for 50 time-steps, where the predicted trajectories are solid lines, the ground truth trajectories are dashed lines and the line colour gets darker along each particle's trajectory. The predicted trajectories were generated by the f\textsc{nri} (leaned) model using the edge-types inferred by the encoder on the prior 50 time-steps (not shown) and the initial state \textit{only} (i.e. predicted steps = 50). In each of these examples the edge-accuracy was 100\%.}
\label{fig:trajectories}%
\end{figure}

Particle trajectories (see figure \ref{fig:trajectories}) are generating by solving Newton's equations of motion using leapfrog integration with a time-step of $1.0 \,$ms. To obtain our training, validation and testing datasets, these trajectories are sub-sampled every 100 time-steps. For each simulated system we generate 50k training examples, 10k validation examples and 10k test examples, where each example contains 100 time-samples with a step size of 0.1$\,$s.

The only major change we make in generating our simulations relative to those generated by Kipf et al. \yrcite{Kipf2018NeuralSystems} is as follows. For each example, rather than randomly connecting each pair of particles by a spring with probability 0.5, the number of springs $n_s \in \lbrace 0, 1, ..., \frac{1}{2}N(N-1) \rbrace$ is drawn from a uniform distribution. Particles are then randomly connected using this number of springs. This means the probability a pair of particles is connected by a spring is still 0.5, while providing a significantly greater variety of interaction graphs. This is desirable as it means a decoder which learns some kind of `average interaction' will perform poorly. Furthermore, when particles are instead randomly connected with probability 0.5, the total number of springs follows a binomial distribution. It is possible the model could learn to use this fact to preferentially assign a number of springs close to the centre of this distribution. This could artificially inflate the obtained edge accuracies and mean that a trained model is less successful when it is used to predict the dynamics of less familiar interaction graphs.

We apply a similar technique when assigning charges. Rather than assigning a charge to each particle with probability 0.5; the number of charges $n_c \in \lbrace 1, ..., N \rbrace$ is drawn from a uniform distribution, then this number of particles are randomly assigned positive charges.

In both of the \textsc{i+c} and \textsc{i+c+f} systems, constants $k_I$, $k_F$, $l$ and $C$ are the same for all interactions and are kept constant between systems (with $k_I = k_F = 0.1$ Nm$^{-1}$, $C = 0.2$ Nm${^2}$ and $l = 1\,$m). The particles of mass 1$\,$Kg, interact in a square 2D box (side-length 5$\,$m, centred on the origin) where their initial locations are sampled from an isotropic 2D Gaussian distribution $\mathcal{N}(0,0.5 \, \text{m})$ and the initial velocity of each particle is a random vector with fixed length 0.5 ms$^{-1}$.


\section{Experimental Details}
\label{S:experimental_section}

In all experiments the models were optimised using the Adam algorithm \cite{Kingma2014Adam:Optimization} with a learning rate of 0.0005, decayed by a factor of 0.5 every 200 epochs. All experiments were run for 500 training epochs using a batch size 128 with shuffling. For \textit{learned} and \textit{true graph} models, checkpointing used the reconstruction loss on the validation set for 10 prediction steps. For \textit{supervised} models, checkpointing used the edge accuracy on the validation set. In the \textsc{nri} and f\textsc{nri} models, the concrete distribution was used with a softmax temperature $\tau = 0.5$.

In the work by Kipf et al. \yrcite{Kipf2018NeuralSystems} edge-types are inferred by observing the trajectories for 50 time-steps of size 0.1$\,$s. These same 50 time-steps are then supplied to decoder for reconstruction. We modify this training routine by supplying the trajectories of the first 50 time-steps to the encoder and the \textit{next} 50 time-steps to the decoder. In order to do this, the simulations we generate are twice as long as the training and validation trajectories used by Kipf et al. This modification is used when training \textit{all} the models in this work.

For all artificial neural networks we use the same architecture and hyperparameters as Kipf et al. \yrcite{Kipf2018NeuralSystems}; using hidden and output dimensions of 256, batch-normalization and \textsc{elu} activations. During training of the decoder, we use an $M$ value of 10, meaning every 10\textsuperscript{th} time-step the decoder receives a ground truth state. We note that in order to prevent exploding gradients in the encoder of the sf\textsc{nri} model when training on the \textsc{i+c} system, a tiny amount of L2 regularisation was added to the loss function (5e-8 for learned, 2e-5 for supervised).

Table \ref{tab:model_sizes} compares the size of the different models in terms of number of parameters and summarises the number of edge-types used in each model in our experiments. These $K$ and $K_a$ values were chosen as for each model they allow for a complete description of the interactions present in each system without redundancy.
\begin{table}[h]
\footnotesize
  \centering
    \caption{\footnotesize Summary of the number of edge-types used by each model (i.e. the dimension $K$ of edge-type vectors $\mathbf{z}_{ij}$) as well as the total number of parameters in the encoder and decoder of each model. \\}
    \label{tab:model_sizes}
    \begin{tabular}{l|ccc|ccc}
      \toprule
      \vspace{1mm}
       & \multicolumn{3}{c}{\ubold{I+C}} & \multicolumn{3}{c}{\ubold{I+C+F}}\\
       & $K$ & \scriptsize Encoder & \scriptsize Decoder & $K$ & \scriptsize Encoder & \scriptsize Decoder \\
       \midrule
       \textsc{nri} & \scriptsize 4 & \scriptsize 710,660 & \scriptsize 406,020 & \scriptsize 8 & \scriptsize 711,688 & \scriptsize 678,404 \\
       f\textsc{nri} & \scriptsize 2+2 & \scriptsize 710,660 & \scriptsize 406,020 & \scriptsize 2+2+2 & \scriptsize 711,174 & \scriptsize 542,212 \\
       sf\textsc{nri} & \scriptsize 2 & \scriptsize 710,146 & \scriptsize 269,828 & \scriptsize 3 & \scriptsize 710,403 & \scriptsize 337,924 \\
      \bottomrule 
    \end{tabular}
\end{table}

In our edge and trajectory prediction experiments, the following baselines are used:
\begin{itemize}
\item \textbf{Supervised}: The encoder is trained in isolation and the ground-truth interaction graphs are provided as labels. For the \textsc{nri} and f\textsc{nri} models we train using the cross-entropy error, and for the sf\textsc{nri} model we use the binary cross-entropy error. All models are trained using a dropout of $p = 0.5$ on the hidden layer representation of every MLP to avoid overfitting, and the edge accuracy on the validation set is used for checkpointing.
\item \textbf{True Graph}: The decoder is trained in isolation and the ground-truth interaction graphs are provided as inputs and we train using the reconstruction error ($M=10$).
\item \textbf{Static}: The decoder copies the previous state vector $\mathbf{x}^{t+1} = \mathbf{x}^t$ for $M$ prediction steps.
\end{itemize}


\section{Edge Accuracy}
\label{edge_section}
In order to calculate the edge accuracies, we have to work out the permutation of the edge-type labels the network uses. For the \textsc{nri} model this is straightforward as the edge-types vectors are already one-hot. For each batch, we compute the edge accuracy for each label permutation. We expect the index permutation which gives us highest edge accuracy to correspond to the permutation the network uses. We can confirm this to be true by looking at the frequency distribution of which label permutations give us this max accuracy over the whole dataset. If the network has settled on a label permutation, we observe all batches to give the max accuracy for the same label permutation. In the f\textsc{nri} and sf\textsc{nri} models where the edge-type vectors are no longer one-hot, this process is more slightly complicated as we also have to account for layer-graph label permutations.

In the results tables, edge accuracies are decomposed into the accuracy for each interaction type. The combined accuracy is calculated such that it only receives a contribution when the predicted edges between a pair of nodes are correct for all interaction types. This gives the combined accuracy a consistent meaning between the models.

\end{document}